\documentclass{main}
\usepackage{bm}
\usepackage{array}
\usepackage{subfigure}
\usepackage{multirow}
\usepackage[ruled,linesnumbered]{algorithm2e}
\usepackage{stfloats}
\title{\vspace{-4cm}CE-based white-box adversarial attacks will not work using super-fitting}

\author{
{Youhuan Yang $^{1,2}$,}
{Lei Sun \xff $^{2}$,}
{Leyu Dai $^{2}$,}
{Song Guo $^{2}$,}
{Xiuqing Mao $^{2}$,}
{Xiaoqin Wang $^{2}$,}
{Bayi Xu $^{1,2}$}
}

\address{{1\quad School of Cyber Science and Engineering, Zhengzhou University, Zhengzhou 450000, China}\\
{2\quad Information Engineering University, Zhengzhou 450001, China}\vspace{3mm}}

\begin{document}

\maketitle

\setcounter{page}{1}

\setlength{\baselineskip}{14pt}

\begin{abstract}
    Deep neural networks are widely used in various fields because 
    of their powerful performance. However, recent studies have shown 
    that deep learning models are vulnerable to adversarial attacks, 
    i.e., adding a slight perturbation to the input will make the model 
    obtain wrong results. This is especially dangerous for some systems 
    with high-security requirements, so this paper proposes a new defense 
    method by using the model super-fitting state to improve the model's 
    adversarial robustness (i.e., the accuracy under adversarial attacks). 
    This paper mathematically proves the effectiveness of super-fitting and 
    enables the model to reach this state quickly by minimizing unrelated 
    category scores (MUCS). Theoretically, super-fitting can resist any 
    existing (even future) CE-based white-box adversarial attacks. 
    In addition, this paper uses a variety of powerful attack algorithms 
    to evaluate the adversarial robustness of super-fitting, and the 
    proposed method is compared with nearly 50 defense models from 
    recent conferences. The experimental results show that the 
    super-fitting method in this paper can make the trained model 
    obtain the highest adversarial robustness.
\end{abstract}


\section{Introduction}

    \noindent Many studies have shown that the deep learning models \cite{1,2,3,4} 
    widely used in various fields are very fragile. Attackers could 
    fool the classifier to achieve evasion attacks by adding slight 
    perturbations to the original examples (adversarial examples) that 
    are difficult to distinguish by human eyes. Meanwhile, various types 
    of defense algorithms with different capabilities (adversarial robustness) 
    have been proposed. Though these algorithms can well handle existing 
    attack methods, they cannot achieve the same effect on future attack 
    methods. Image preprocessing \cite{5,6} and defensive distillation \cite{7} 
    do not depend on adversarial examples. These algorithms are effective 
    for various attack methods, but their performance is suboptimal. 
    Adversarial training \cite{8} is considered the most effective defense 
    method at present (many methods are based on adversarial training), 
    and these algorithms rely on adversarial examples (or specific types 
    of adversarial attack algorithms). The existing adversarial defense 
    methods share some shortcomings: 1) the adversarial robustness (accuracy) 
    is too low, i.e., the performance of the defense model on adversarial 
    examples is much lower than that on the original clean examples; 2) the 
    generalization against unknown attacks is poor, and these algorithms 
    often cannot defense effectively when a more advanced (or future) 
    attack algorithm is proposed; 3) the implementation is complicated, 
    and most of these defense methods have too many hyperparameters. 
    Also, the hyperparameters of the training process requires careful 
    tuning. Therefore, it is acknowledged that adversarial defense is a 
    very difficult task.

    Since Szegedy et al. \cite{9} first discovered the existence of 
    adversarial examples, various studies on adversarial attack 
    algorithms have developed rapidly. However, most of these 
    attack algorithms (including white- and black-box attacks) are 
    based on Cross-Entropy (CE) loss, and only a few attack algorithms 
    use other (or self-define) loss functions (such as AutoAttack \cite{10}, 
    C\&W \cite{11}, etc.). This paper proposes a new defense method based on 
    super-fitting for these widely CE-based white-box adversarial 
    attacks. Super-fitting is a state of model training. In this state, 
    the gradient information calculated by the CE loss will die out (i.e., 
    Gradient Vanishing). In the experiment, it is found that the model in 
    this state defense against almost any CE-based white-box adversarial 
    attacks in existence. In the subsequent introductions, this paper 
    explains the reason for this phenomenon and the mathematical principle 
    of defense and proposes a method called MUCS to enable the defense 
    model to quickly reach the super-fitting state. The contributions of 
    this paper are as follows:

\begin{itemize}

    \item This paper proposes the concept of super-fitting in the model 
    training phase. In this state, the model can defend almost all 
    existing CE-based white-box adversarial attacks, and the training 
    phase is not related to the adversarial attack algorithm. Through analysis, 
    it can be seen that the model trained using the super-fitting method can 
    defend against future CE-based white-box adversarial attacks.

    \item This paper analyzes the reason for the super-fitting phenomenon 
    and proposes the MUCS method to reduce the number of model training 
    iterations to reach this state, which makes it easy to train a super-fitting 
    model without high-power graphics.

    \item Finally, this paper experimentally proves the effectiveness of 
    the proposed method. Compared with nearly 50 defense models from top 
    conferences (e.g., ICML, NeurIPS, ICLR, ICCV, CVPR). The experimental 
    results indicate that super-fitting obtains the highest adversarial 
    robustness against CE-based white-box adversarial attacks.

\end{itemize}

\section{Preliminaries}

\subsection{Adversarial attack}

    \noindent Although deep neural networks perform well in many fields 
    (e.g., face recognition, object detection, medical image process, 
    malicious traffic detection, etc.), many studies \cite{8,12,13,14} show 
    that deep neural networks are very fragile and vulnerable to adversarial 
    attacks, i.e., adding slight perturbations to the input that are hard 
    to distinguish with the human eyes (adversarial perturbations) will make 
    the classifier obtain unexpected results. The constrained optimization 
    problem is defined as: 

        \begin{equation}
            y\neq\arg\max f(x+\delta)
        \end{equation}
        \begin{equation*}
            s.t.\ {||\delta||}_p\le\epsilon
        \end{equation*}

    In Eq. (1), ${||\bullet||}_p$ denotes $L-p$ norm. $p\in\{1,2,\infty\}$, and 
    this paper sets $p=\infty$, i.e., just considers $L-\infty$ attack; $\epsilon$ 
    is a hyperparameter that limits the size of adversarial perturbations; 
    $x$ and $x+\delta\in[0,1]^{D}$ denote origin images and adversarial images, 
    respectively; $\delta\in[-\epsilon,+\epsilon]^{D}$ denote perturbations; $x$ is the input of 
    classifier $f$, and $y$ is the true label. Goodfellow et al. \cite{12} proposed 
    the Fast Gradient Sign Method (FGSM) based on the definition of 
    adversarial attacks in Eq. (1). An adversarial attack could be 
    formulated as follows:
        \begin{equation}
            x_{adv}=x+\epsilon sign(\mathrm{\nabla}_x\mathcal{L}(f(x),y))
        \end{equation}

    In Eq. (2), $sign$ represents the sign function, $\mathcal{L}$ is the loss 
    function (i.e., CE loss in this paper), and $x_{adv}$ is the adversarial 
    example generated by the adversarial attack algorithm. Based on FGSM, 
    Kurakin et al. \cite{13} proposed a stronger multi-step iterative attack algorithm 
    called Basic Iterative Method (BIM), which uses a smaller step size in 
    each gradient ascent iteration process (attack process). Similar to the 
    BIM algorithm, Madry et al. \cite{8} found that adding random noise to the input 
    during the restart phase of the attack algorithm can make it stronger (i.e., 
    approaching the lower bound of robustness). Projected Gradient Descent (PGD) 
    generates an adversarial example by performing the iterative update:
        \begin{equation}
            x_{adv}^{t+1}={{Proj}_{(x,\epsilon)}(x}_{adv}^t+\alpha{sign(\mathrm{\nabla}}_{x_{adv}^t}\mathcal{L}(f(x_{adv}^t),y)))
        \end{equation}
    where,
        \begin{equation*}
            x_{adv}^0=x_{orig}+\delta_0
        \end{equation*}
        \begin{equation*}
            s.t.\ {||\delta_0||}_\infty\le\epsilon
        \end{equation*}

    In Eq. (3), $t$ is the number of iterations, $x_{orig}$ is the original image, $\delta_0$ 
    is the randomly initialized noise, ${Proj}_{(x,\epsilon)}$ is the operation to 
    clip perturbations in $\epsilon$-ball constrain, and $\alpha$ is the step size 
    of every single attack. Many researchers have made improvements to PGD. 
    For example, Croce et al. \cite{10} proposed Adaptive Projected Gradient 
    Descent (APGD). Unlike the original PGD, APGD does not have a fixed 
    step size, i.e., $\alpha$ will dynamically change in each iteration according 
    to the value of the loss function.

    Obtaining an attack starting point $x_{adv}^0$ by random initialization is always 
    suboptimal (e.g., ODI \cite{15} and AutoAttack \cite{10}). So, Liu et al. \cite{16} proposed 
    Adaptive Auto Attack ($A^3$), which is quite different from PGD in starting point 
    initialization. To be specific, given a random direction of diversification $w_d$ 
    sampled from uniform distributions $U{(-1,1)}^K$ ($K$ is the total number of 
    categories for the classification task), $A^3$ firstly calculates a normalized 
    perturbation vector as follows:
        \begin{equation}
            v(x,f,w_d)=\frac{\mathrm{\nabla}_xw_d^Tf(x)}{||\mathrm{\nabla}_xw_d^Tf(x)||}
        \end{equation}
    
    Then, the starting point is calculated through the following iterative process:
        \begin{equation}
            x_{init}^{t+1}={P_{\left(x,\epsilon\right)}(x}_{init}^t{+\alpha}_{init}sign(v(x_{init}^t,f,w_d)))
        \end{equation}
    
    When the $T$-th initialization iteration ends, $x_{adv}^0=x_{init}^T$ is 
    selected as the attack starting point, and then $A^3$ chooses an 
    attack process similar to PGD. Differently, its step size is 
    not fixed compared to PGD. The step size of the $n$-th round of 
    attack (assuming a total of $N$ rounds of iterative attacks) 
    is calculated as follows:

        \begin{equation}
            \alpha^n=\frac{1}{2}\epsilon\bullet(1+cos(\frac{n\ mod\ N}{N}\ \pi))
        \end{equation}

    It is worth noting that the original $A^3$ uses the statistical results 
    to modify the distribution information of the $w_d$ part of the dimension 
    after randomly obtaining $w_d$. The subsequent experimental part of 
    this paper does not take this approach. This is because the increase 
    in revenue is small. Also, the original $A^3$ uses Margin loss for attacks, 
    but this paper uses the CE loss.

    It can be seen that in general, the white-box attack is based on the 
    gradient, and most attack methods use the CE loss when calculating 
    the gradient information to attack the classifier. The main reason 
    is that the current classification tasks generally use the CE loss 
    for model training. Its calculation formula is as follows:
        \begin{equation}
            \mathcal{L}(z,y)=-z_y+ln{(\sum_{i=0}^{K-1}e^{z_i})}
        \end{equation}

    In Eq. (7), $z$ is the output of the classifier (logits vector, $z\in\mathbb{R}^K$), 
    and $y$ is the true label of the example ($y\in[0,K)$). Although CE loss is 
    the most popular loss function to obtain gradients for adversarial attacks, 
    its shortcomings are also obvious. When $z_k\rightarrow-\infty$ ($k\in[0,K)$, 
    $k\neq y$),there will be a \textit{Gradient Vanishing} situation. The specific explanation and 
    formulation will be presented in Section 3.

\subsection{Adversarial defense}

    \noindent To address the adversarial attack threat, different types of 
    defense algorithms have been proposed, e.g., image pre-processing \cite{5,6}, 
    adversarial training \cite{8}, gradient obfuscation/hiding \cite{17}, 
    regularization \cite{18,19}, etc. Among them, adversarial training is 
    currently considered to be the most effective defense method, and 
    its optimization process is defined as follows:
        \begin{equation}
            \min_{\theta}E_{(x,y)^{D}}\max_{{||\delta||}_p\le\epsilon}\mathcal{L}(\theta,f(x+\delta),y)
        \end{equation}

    Eq. (8) contains two optimization problems, namely the inner 
    maximization and outer minimization problems. The inner maximization 
    problem always tries to find the worst-case examples (adversarial examples), 
    and the outer minimization problem is to train a model robust to adversarial 
    examples. Many studies are based on the above-mentioned standard adversarial 
    training. For example, CAT \cite{20} uses a small number of iterations in the 
    initial stage of model training, and it increases the number of iterations 
    of the attack when the model achieves a high accuracy against the current 
    attack. FAT \cite{21} adopts early stopping in the inner maximization problem 
    to approach the lower robustness boundary. In addition to this, there is 
    a part of work dedicated to reducing the time of adversarial training, 
    e.g., YOPO \cite{22}, FAST-AT, \cite{23}, etc. For example, TRADES \cite{18} attempts 
    to balance accuracy between clean examples and adversarial examples. 
    There are many methods based on adversarial training, and the performance 
    gap between these methods is not large. A common point of view is the 
    improvement of robustness largely depends on the internal maximization 
    problem. A stronger attack algorithm will make the defense model more 
    robust, but the cost will also be greater.

    Different from adversarial training, defensive distillation \cite{7} is 
    a gradient hiding method, which uses real labels (hard labels) to 
    train the teacher network and uses the output of the teacher network 
    as the soft label of the student network. The calculation of the 
    softmax function with temperature (used to calculate the loss value) 
    is defined as:
        \begin{equation}
            F(z,y,T)=\frac{e^{z/T}}{\sum_{i=0}^{K-1}e^{z_i/T}}
        \end{equation}

    Particularly, Eq. (9) is a normal softmax function when $T=1$. It can be seen 
    that although defensive distillation can hide the gradient information of 
    the teacher network to a certain extent, gradient information will still 
    be leaked through the student network.
    
    Although various defense methods have been proposed to resist 
    adversarial attacks, the results are unsatisfactory, and there is 
    no effective method that can make the defense model achieve a similar 
    accuracy on adversarial examples to that on clean examples. Most 
    importantly, current defense methods cannot defend well both mathematically 
    and experimentally, even in the face of CE-based adversarial attacks that 
    have existed for a long time.
        
\section{Super-fitting}

    \noindent In this section, the super-fitting phenomenon in the 
    model training phase and its explanation are first introduced. 
    Then, the method MUCS is proposed to accelerate the model to reach 
    the super-fitting state.

\subsection{Phenomenon}

    \noindent The general model training state changes from under-fitting 
    (at the beginning of the training phase) to fitting (the training is 
    completed) and finally reaches the state of over-fitting. Since over-fitting 
    will reduce the generalization of the trained model, researchers always 
    choose to stop training when the model reaches the fitting state during 
    the training process. However, through an experiment, this paper found 
    that the concepts of different fitting states are relative, and over-fitting 
    on the original/clean examples is under-fitting on the adversarial examples. 
    By increasing the number of iterations of the training phase, the model can 
    also reach the state of fitting on adversarial examples. This state is far 
    behind over-fitting for clean examples, which is called super-fitting in 
    this paper. This paper uses MiddleCNN (the network consists of three 
    convolutional layers with 64, 128 and 256 output channels, respectively, 
    and two fully connected layers with 1024 and 10 units, respectively. Each 
    convolutional layer is followed by BatchNormalization \cite{24}, LeakReLu and 
    Maxpooling layer.) and ResNet-18 \cite{1} to train defense models on the CIFAR-10 
    dataset to make them super-fitting. Fig. 1 shows the partial information of 
    the two defense models in the training phase (top MiddleCNN, bottom ResNet-18), 
    including the accuracy of clean examples for the training set and the accuracy 
    of adversarial examples for the test set (left) and the value of the loss 
    function (right) calculated on adversarial examples using the CE loss.
        \begin{figure*}[!ht]
            \centering
            \includegraphics[width=0.9\textwidth]{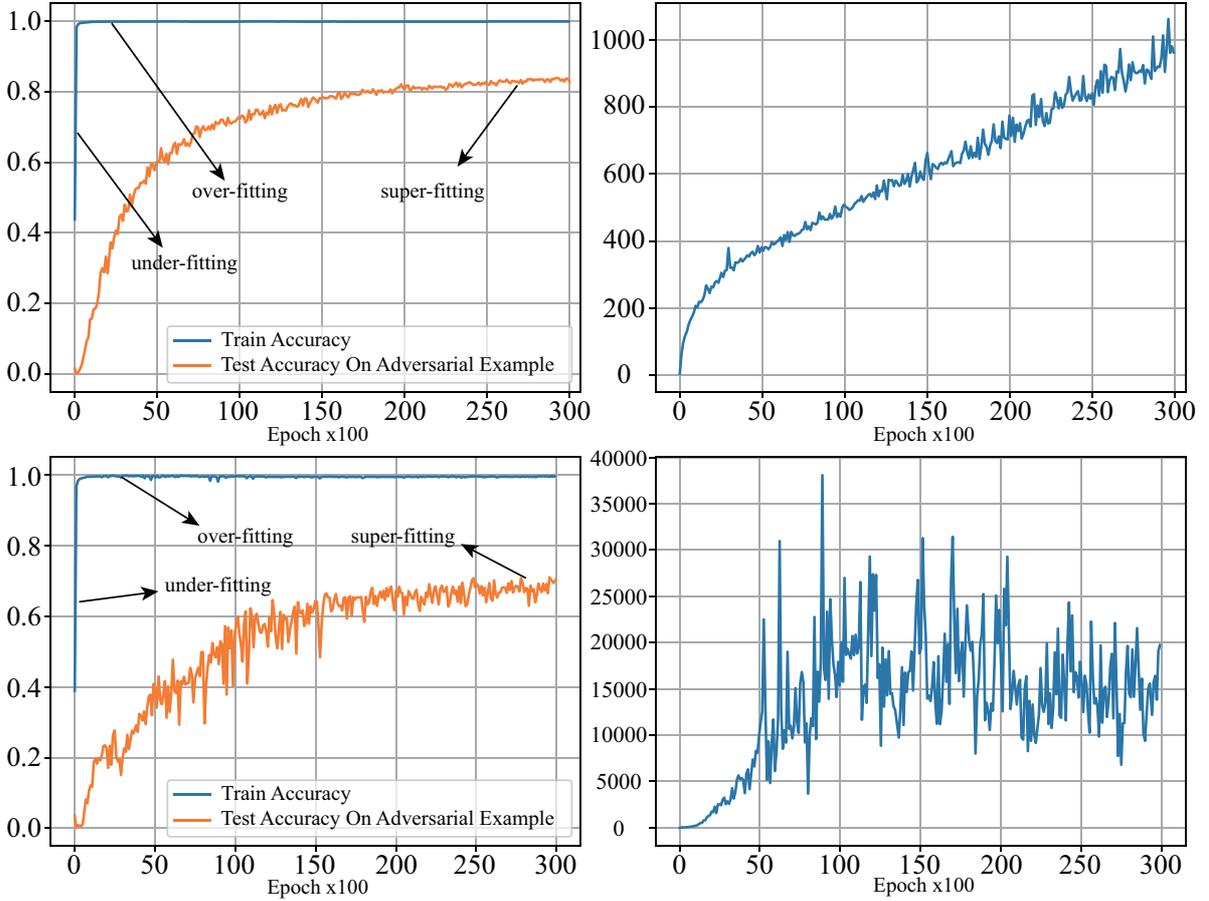}
            \vspace{-1em}
            \caption{Different fitting states on CIFAR-10}
            \label{Fig01}
        \end{figure*}

    In the experiment, both models are trained using clean examples, the CE loss, and 
    the Adam \cite{25} optimizer (the learning rate is set to 1e-3, and other parameters 
    used in the training process are set to the default). No defensive measures 
    are used, and just the number of training iterations is increased (max epoch 
    is 30000). Afterward, PGD-20 ($\epsilon=8.0/255$, step size is $\epsilon/10$) is 
    used on the CIFAR-10 test set to evaluate the accuracy. The accuracy of MiddleCNN 
    on the adversarial examples generated by PGD-20 is 83.72\% (the accuracy 
    on clean examples is 88.97\%), and the adversarial robustness of ResNet-18 
    is 68.75\% (the accuracy on clean examples is 83.85\%).

    Obviously, super-fitting only sacrifices a part of generalization (the 
    accuracy on the clean examples is lower than that in the fitting state), 
    but it greatly improves the adversarial robustness of the model. Surprisingly, 
    the loss function value of the adversarial examples calculated with the CE loss 
    in Fig. 1 is increasing (theoretically it should be a very small value). To 
    find the reason for this phenomenon, this paper shows the statistics of the 
    prediction results (logits) of each category of CIFAR-10 by MiddleCNN in the 
    super-fitting state in Fig. 2.
        \begin{figure*}[!ht]
            \centering
            \includegraphics[width=0.9\textwidth]{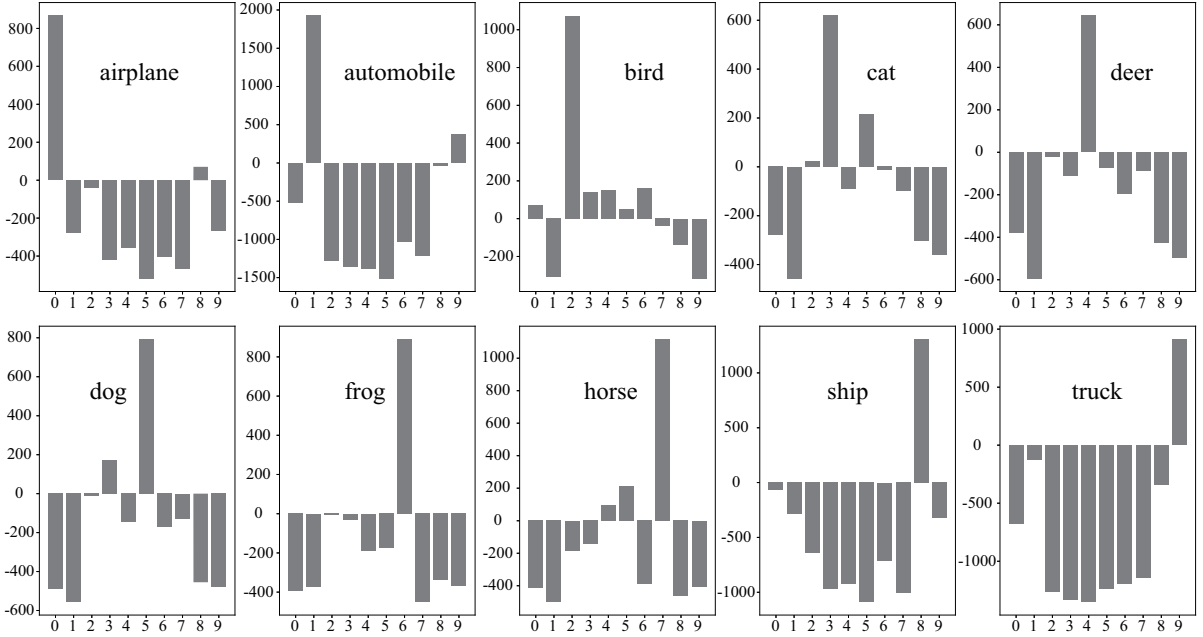}
            \vspace{-1em}
            \caption{Logits statistics for each category}
            \label{Fig02}
        \end{figure*}
    
    In Fig. 2, the output information of the same categories of examples 
    is in the same subfigure. There are 10 bars in each subplot, and the 
    values are the mean of model output in each dimension on the category 
    of examples, i.e., the prediction scores of each category given by the 
    model. It can be seen from the figure that the prediction scores of the 
    examples in the super-fitting state of the MiddleCNN are polarized. The 
    predicted values of examples in the dimension of the real label are much 
    larger than others, and the values in the dimensions that do not belong 
    to the real label are generally much smaller than 0.

    Inspired by the distribution of the example output results in Fig. 2, 
    this paper analyzes the impact of logits changes on the CE loss. 
    It is found that when the model outputs meet $z_k\rightarrow-\infty$(
    $k\in[0,K)\ and\ k\neq y$), the CE-based white-box adversarial 
    attack algorithm will not work.

\subsection{Analysis}

    \noindent Since $\ln{x}$ cannot be differentiated when $x\le0$, the CE loss shown 
    in Eq. (7) often adopts an equivalent calculation:
        \begin{equation}
            \mathcal{L}(b,t)=-\sum_{i=0}^{K-1}{t_ib}_i
        \end{equation}
    where,
        \begin{equation*}
            t_i=\left\{\begin{matrix}
                1&i=y\\ 
                0&others
            \end{matrix}\right.
        \end{equation*}
        \begin{equation}
            b_i=ln{\frac{e^{z_i}}{\sum_{j=0}^{K-1}e^{z_j}}=z_i-ln{\sum_{j=0}^{K-1}e^{z_j}}}
        \end{equation}
    $b$ is the result of the model output after passing through the softmax 
    and ln functions; $t$ is the one-hot encoding format of the true label 
    of the example. From Eq. (10) and Eq. (11), it can be deduced:
        \begin{equation}
            \frac{\partial\mathcal{L}}{\partial b_i}=-t_i=\left\{\begin{matrix}
                -1&i=y\\ 
                0&others
            \end{matrix}\right.
        \end{equation}
        \begin{equation}
            \frac{\partial\mathcal{L}}{\partial z_i}=\sum_{j=0}^{K-1}{\frac{\partial\mathcal{L}}{\partial b_j}\frac{\partial b_j}{\partial z_i}=\frac{\partial\mathcal{L}}{\partial b_y}\frac{\partial b_y}{\partial z_i}=-\frac{\partial b_y}{\partial z_i}}
        \end{equation}
        \begin{equation}
            b_y=z_y-ln{\sum_{j=0}^{K-1}e^{z_j}}
        \end{equation}

        Then,
        \begin{equation}
            \frac{\partial\mathcal{L}}{\partial z_i}=\left\{\begin{matrix}
                \frac{e^{z_y}}{\sum_{j=0}^{K-1}e^{z_j}}-1&i=y\\ 
                \frac{e^{z_i}}{\sum_{j=0}^{K-1}e^{z_j}}&others
            \end{matrix}\right.
        \end{equation}

    It is easy to find $e^{z_k}\rightarrow0$ when $z_k\rightarrow-\infty(k\in[0,K)\ and\ k\neq y)$. Then,
        \begin{equation}
            \frac{\partial\mathcal{L}}{\partial z_i}=0
        \end{equation}

    There will be a gradient vanishing at this time, i.e., the CE-based 
    white-box adversarial attack will not work because the gradient 
    information cannot be propagated. Therefore, this paper mathematically 
    explains how super-fitting resists adversarial attacks. Based on this, 
    the method proposed in this paper can defend against future CE-based 
    white-box adversarial attacks.

    However, super-fitting will face two problems: 1) $z_k$ cannot reach the 
    mathematical $-\infty$; 2) cost too much time (ResNet-18 needs about 
    30,000 iterations or a few days to reach the super-fitting state 
    even with an RTX3090 graphics card). Under the floating-point 
    operation of modern computers, the first problem can be 
    solved easily by optimizing $z_k$ to a very small value ($z_k<0$, 
    it does not need to reach the mathematical $-\infty$). In this 
    case, $e^{z_k}$ will be zero, and Eq. (16) holds. To solve 
    the second problem, a new optimization process is proposed 
    in this paper to minimize unrelated category scores, and the 
    specific implementation is introduced in Section 3.3.

\subsection{MUCS: Minimize unrelated category scores}

    \noindent Although training the model with the CE loss can also 
    make it super-fitting, the process consumes a lot of time. Also, 
    it is difficult to achieve a perfect super-fitting state (e.g., 
    experiment in Fig. 1) when the model parameters are large (e.g., 
    ResNet-18). To address these issues, this paper proposes a new 
    optimization method called MUCS, and the new optimization 
    problem is defined as:
        \begin{equation}
            minimize\ {(z}_s-z_y)
        \end{equation}
    where,
        \begin{equation*}
            s=argmax\ z_i \ and \ s\neq y
        \end{equation*}
    
    In Eq. (17), $z_s$ is the maximum value of prediction scores other than the 
    true label (i.e., class scores that are not related to the true label). In 
    fact, the goal of minimizing the scores of all irrelevant classes can be achieved 
    just by minimizing $z_s$.
    
    Table 1 shows the effect of MUCS and the CE loss on the super-fitting state 
    of CIFAR-10, including the convergence speed and the final robustness 
    accuracy (against PGD-20).

\doublerulesep 0.1pt
\begin{table*}
\centering
\begin{footnotesize}
\caption{Performance of MUCS compare to the CE loss} 
\begin{tabular}{ccccc}
\hline\hline
\multicolumn{2}{c}{Model}&Iterations&Clean Accuracy&Robustness Accuracy\\
\hline
\multirow{2}{*}{MiddleCNN}&CE-Loss&30,000&88.97\%&83.45\%\\
&MUCS&500&86.86\%&85.83\%\\
\multirow{2}{*}{ResNet-18}&CE-Loss&30,000&84.10\%&69.72\%\\
&MUCS&3,000&82.65\%&84.21\%\\
\hline\hline
\end{tabular}
\end{footnotesize}
\end{table*}

\doublerulesep 0.1pt
\begin{table*}
\centering
\begin{footnotesize}
\caption{Performance comparison of super-fitting and defensive distillation (DD)} 
\begin{tabular}{p{0.15\textwidth}<{\centering} p{0.15\textwidth}<{\centering} p{0.1\textwidth}<{\centering} p{0.1\textwidth}<{\centering} p{0.1\textwidth}<{\centering}}
\hline\hline
\multicolumn{2}{c}{Model}&Clean&FGSM&BIM-20\\
\hline
\multirow{2}{*}{MiddleCNN}&DD&86.79\%&82.31\%&82.21\%\\
&Ours&89.05\%&89.05\%&89.05\%\\
\multirow{2}{*}{ResNet-18}&DD&80.98\%&78.40\%&78.35\%\\
&Ours&83.87\%&83.87\%&83.87\%\\
\hline\hline
\end{tabular}
\end{footnotesize}
\end{table*}

\doublerulesep 0.1pt
\begin{table*}
\centering
\begin{footnotesize}
\caption{Performance of super-fitting against powerful attack}

\begin{tabular}{p{0.2\textwidth}<{\centering} p{0.2\textwidth}<{\centering} p{0.1\textwidth}<{\centering} p{0.1\textwidth}<{\centering} p{0.1\textwidth}<{\centering} p{0.1\textwidth}<{\centering}}
\hline\hline

\multicolumn{2}{c}{CIFAR-10 Defense Model}&Clean&PGD&APGD&$A^3$\\
\hline

MiddleCNN\textbf{(20.33MB)}&super-fitting&87.32\%&\textbf{86.56\%}&\textbf{84.33\%}&\textbf{78.59\%}\\
\hline

\multirow{7}{*}{ResNet-18(43.96MB)}&Adv-regular\cite{19}&\textbf{88.99\%}&80.76\%&68.03\%&69.29\%\\
&FBTF\cite{23}&86.74\%&69.41\%&60.67\%&47.26\%\\
&CNL\cite{26}\ddag&84.60\%&71.25\%&64.62\%&76.71\%\\
&Understanding FAST\cite{27}&83.97\%&66.19\%&60.77\%&47.34\%\\
&Proxy\cite{28}&84.01\%&65.09\%&60.49\%&59.68\%\\
&DNR\cite{29}&84.35\%&62.12\%&58.03\%&45.62\%\\
&OAAT\cite{30}&83.95\%&61.44\%&57.78\%&57.34\%\\
\hline

WRN-28-4(48.95MB)&MMA\cite{31}&87.08\%&61.89\%&60.71\%&52.03\%\\
\hline

ResNet-50(95.66MB)&Robustness\cite{32}&87.06\%&61.67\%&60.50\%&52.57\%\\
\hline

\multirow{11}{*}{WRN-28-10(205.17MB)}&MART\cite{33}\dag&84.35\%&61.32\%&57.99\%&61.01\%\\
&Feature-scatter\cite{34}&84.92\%&62.29\%&58.69\%&72.34\%\\
&Adv-inter\cite{35}&85.51\%&63.51\%&60.06\%&75.23\%\\
&AWP\cite{36}\dag&85.86\%&63.69\%&60.51\%&65.39\%\\
&Geometry\cite{37}\dag\ddag&86.15\%&63.96\%&60.89\%&67.81\%\\
&Hydra\cite{38}\dag&86.28\%&63.50\%&60.61\%&58.98\%\\
&Pre-train\cite{39}&86.32\%&63.12\%&60.39\%&57.65\%\\
&Rst\cite{40}\dag&86.54\%&63.13\%&60.53\%&63.51\%\\
&ULAT\cite{41}\dag&86.71\%&63.32\%&60.86\%&67.42\%\\
&Fix-data\cite{42}&86.80\%&63.53\%&61.21\%&66.48\%\\
&RLPE\cite{43}\dag&87.03\%&63.60\%&61.38\%&65.23\%\\
\hline

\multirow{10}{*}{WRN-34-10(257.42MB)}&TRADES\cite{18}\ddag&86.95\%&63.25\%&61.13\%&57.57\%\\
&AWP\cite{36}&86.95\%&63.17\%&61.15\%&60.85\%\\
&Proxy\cite{28}\dag&86.92\%&63.09\%&61.15\%&61.48\%\\
&Self-adaptive\cite{44}\ddag&86.74\%&62.80\%&60.93\%&56.71\%\\
&Sensible\cite{45}&86.96\%&62.67\%&60.74\%&62.26\%\\
&YOPO\cite{22}&86.94\%&62.03\%&60.17\%&46.09\%\\
&IAR/SAT\cite{46}&86.92\%&61.72\%&59.90\%&53.82\%\\
&LBGAT\cite{47}\ddag&86.94\%&61.37\%&59.60\%&54.84\%\\
&FAT\cite{21}&86.90\%&61.25\%&59.54\%&57.42\%\\
&OAAT\cite{30}&86.85\%&61.41\%&59.74\%&65.07\%\\
\hline

WRN-34-15(517.84MB)&RLPE\cite{43}\dag&87.02\%&61.67\%&60.52\%&61.25\%\\
\hline

\multirow{4}{*}{WRN-34-20(866.20MB)}&Hyper-embe\cite{48}&86.81\%&61.44\%&59.79\%&63.12\%\\
&Overfit\cite{49}&86.75\%&61.34\%&59.74\%&58.67\%\\
&ULAT\cite{41}&86.71\%&61.25\%&59.68\%&58.75\%\\
&LBGAT\cite{47}\ddag&86.80\%&61.09\%&59.56\%&58.12\%\\
\hline

\multirow{3}{*}{WRN-70-16(1294.64MB)}&ULAT\cite{41}&87.15\%&61.84\%&60.56\%&60.46\%\\
&ULAT\cite{41}\dag&87.30\%&61.84\%&60.42\%&67.96\%\\
&Fix-data\cite{42}&87.16\%&62.17\%&60.98\%&67.03\%\\

\hline\hline



\multicolumn{2}{c}{CIFAR-100 Defense Model}&Clean&PGD&APGD&$A^3$\\
\hline

MiddleCNN\textbf{(20.68MB)}&super-fitting&59.21\%&\textbf{55.19\%}&\textbf{52.32\%}&\textbf{41.71\%}\\
\hline

\multirow{2}{*}{ResNet-18(44.14MB)}&Overfit\cite{49}&55.16\%&37.95\%&36.30\%&20.46\%\\
&OAAT\cite{30}&57.06\%&36.04\%&34.78\%&32.96\%\\
\hline

\multirow{2}{*}{WRN-28-10(205.39MB)}&Pre-train\cite{39}&58.32\%&36.07\%&35.03\%&35.85\%\\
&Fix-data\cite{42}&59.13\%&36.27\%&35.37\%&37.73\%\\
\hline

\multirow{4}{*}{WRN-34-10(257.64MB)}&AWP\cite{36}&59.23\%&36.21\%&35.42\%&35.54\%\\
&IAR/SAT\cite{46}&59.65\%&34.69\%&33.86\%&26.40\%\\
&LBGAT\cite{47}\ddag&59.86\%&34.76\%&34.01\%&36.56\%\\
&OAAT\cite{30}&60.27\%&34.78\%&34.05\%&34.68\%\\
\hline

WRN-34-20(866.64MB)&LBGAT\cite{47}\ddag&60.47\%&34.81\%&34.07\%&35.23\%\\
\hline

\multirow{3}{*}{WRN-70-16(1295.00MB)}&ULAT\cite{41}&60.44\%&34.12\%&33.54\%&32.50\%\\
&ULAT\cite{41}\dag&62.61\%&35.63\%&35.16\%&40.00\%\\
&Fix-data\cite{42}&\textbf{62.93\%}&36.27\%&35.81\%&39.53\%\\

\hline\hline
\end{tabular}
\end{footnotesize}
\end{table*}

    The experimental parameter settings in Table 1 are consistent with those 
    in Fig. 1. The experimental results show that compared with the CE loss, 
    although the accuracy of MUCS on the original examples (Clean Accuracy) 
    is reduced, its adversarial robustness (accuracy against adversarial 
    attacks) is improved (Robustness Accuracy). Also, the number of iterations 
    required to reach the super-fitting state is greatly reduced (Iterations). 
    Therefore, this paper combines the CE loss and MUCS in the subsequent 
    experiments and uses Eq. (18) to optimize the model:
        \begin{equation}
            minimize\ CE(z,y)\ +\ {(z}_s-z_y)
        \end{equation}

    Eq. (18) can ensure that the number of iterations of model training 
    is greatly reduced, and the accuracy of the original examples and the 
    adversarial examples can reach a high level (i.e., the trade-off 
    will be better).

\section{Experiment}
    
    \noindent In this section, to demonstrate the effectiveness of 
    the method proposed in this paper, super-fitting is compared with 
    defensive distillation, which also uses gradient-based defense. 
    This paper conducts extensive experiments to compare the adversarial 
    robustness of 50 defense models on CIFAR-10 and the more difficult 
    task CIFAR-100. There are 38 CIFAR-10 defense models and 12 CIFAR-100 
    defense models, and most of these are from recent top conferences (e.g., 
    ICML, NeurIPS, ICLR, ICCV, CVPR).

    \textbf{Setup.} 1) In the comparison experiment of super-fitting and defensive distillation, 
    this paper adopts the weak attack algorithms FGSM and BIM-20 for 
    adversarial robustness evaluation; 2) To compare with state-of-the-art 
    defense models, this paper adopts the more powerful PGD-100, APGD-100, 
    and standard $A^3$; 3) The Adam optimizer is used for super-fitting training 
    in all experiments, and the learning rate is 1e-3. The hyperparameters of 
    the attack algorithm are set as $\epsilon=8.0/255$ and step size = $\epsilon/10$. 
    The CE loss is used to calculate the gradient during the attack, and other 
    parameters are kept at default.

    \textbf{Results.} This paper shows the final experimental comparison 
    results in Table 2 and Table 3. Table 2 shows that super-fitting performs 
    better for the gradient-based defense methods. It is theoretically possible 
    to make the gradient vanish completely. Table 3 shows that the model is 
    robust against CE-based white-box adversarial attacks on the CIFAR-10 
    and CIFAR-100 datasets, and the super-fitting method performs better 
    than other methods.
        
    The experiments in Table 2 are performed on CIFAR-10, and two network 
    architectures are taken for comparison (MiddleCNN and ResNet-18). 
    The teacher network and student network model architectures used in 
    defense distillation are consistent, and the distillation temperature 
    is set to 100. The experimental results show that our proposed method 
    works better than defensive distillation. Also, super-fitting can 
    completely defend against weak attack algorithms like FGSM and BIM-20 
    (adversarial accuracy equal to the clean one).

    This paper compares the performance of super-fitting with other recent 
    excellent defense methods on the test sets of CIFAR-10 and CIFAR-100 
    datasets (1280 images are randomly selected, and the batch size is set to 128). 
    Meanwhile, this paper uses the original examples, the adversarial examples 
    generated by PGD-100, APGD-100, and the default version of $A^3$ in nine network 
    model architectures (three basic architectures of MiddleCNN, ResNet, Wide 
    ResNet/WRN \cite{4}) to evaluate the adversarial robustness of the models. The 
    results are shown in Table 3, where the first column presents the defense 
    methods, network architectures, and model parameters (the one marked with \ddag \ denotes 
    that $\epsilon=0.031$, and the models marked with \dag \ denote that they 
    use an additional unlabeled dataset in the training phase). “Clean” presents 
    the accuracy of the model under the original examples; PGD, APGD, and $A^3$ 
    respectively present the accuracy of the models under adversarial examples 
    generated by PGD-100, APGD-100, and the default version of $A^3$ algorithms. 
    Obviously, the super-fitting method proposed in this paper can obtain higher 
    adversarial robustness with smaller models (i.e., fewer parameters).

\section{Conclusion}
    
    \noindent This paper first introduces the super-fitting state in the model 
    training process. In this state, the accuracy of a trained model under 
    clean examples will decrease slightly, but its adversarial robustness 
    will be greatly improved. Then, this paper analyzes the mathematical 
    reasons that lead to the super-fitting phenomenon and theoretically proves 
    that under the floating-point operation of the modern computer, 
    super-fitting can perfectly handle CE-based white-box adversarial 
    attacks (even future ones). Finally, for the problem that super-fitting 
    requires a lot of iterative training, this paper proposes an optimization 
    strategy called MUCS, which greatly reduces the number of iterations required 
    for the model to reach the super-fitting state. Compared with about 50 recent 
    defense models, the proposed defense method performs the best against CE-based 
    white-box adversarial attacks.


\begin{thebibliography}{99}
    \bibitem{1}He, K., et al. Deep residual learning for image recognition. in Proceedings of the IEEE conference on computer vision and pattern recognition. 2016.
    \bibitem{2}Huang, G., et al. Densely connected convolutional networks. in Proceedings of the IEEE conference on computer vision and pattern recognition. 2017.
    \bibitem{3}Szegedy, C., et al. Rethinking the inception architecture for computer vision. in Proceedings of the IEEE conference on computer vision and pattern recognition. 2016.
    \bibitem{4}Zagoruyko, S. and N. Komodakis, Wide residual networks. arXiv preprint arXiv:1605.07146, 2016.
    \bibitem{5}Gu, S. and L. Rigazio, Towards deep neural network architectures robust to adversarial examples. arXiv preprint arXiv:1412.5068, 2014.
    \bibitem{6}Osadchy, M., et al., No bot expects the DeepCAPTCHA! Introducing immutable adversarial examples, with applications to CAPTCHA generation. IEEE Transactions on Information Forensics and Security, 2017. 12(11): p. 2640-2653.
    \bibitem{7}Papernot, N., et al. Distillation as a defense to adversarial perturbations against deep neural networks. in 2016 IEEE symposium on security and privacy (SP). 2016. IEEE.
    \bibitem{8}Madry, A., et al., Towards deep learning models resistant to adversarial attacks. arXiv preprint arXiv:1706.06083, 2017.
    \bibitem{9}Szegedy, C., et al., Intriguing properties of neural networks. arXiv preprint arXiv:1312.6199, 2013.
    \bibitem{10}Croce, F. and M. Hein. Reliable evaluation of adversarial robustness with an ensemble of diverse parameter-free attacks. in International conference on machine learning. 2020. PMLR.
    \bibitem{11}Carlini, N. and D. Wagner, Towards Evaluating the Robustness of Neural Networks, in 2017 IEEE Symposium on Security and Privacy (SP). 2017. p. 39-57.
    \bibitem{12}Goodfellow, I.J., J. Shlens, and C. Szegedy, Explaining and harnessing adversarial examples. arXiv preprint arXiv:1412.6572, 2014.
    \bibitem{13}Kurakin, A., I. Goodfellow, and S. Bengio, Adversarial examples in the physical world. 2016.
    \bibitem{14}Moosavi-Dezfooli, S.-M., A. Fawzi, and P. Frossard. Deepfool: a simple and accurate method to fool deep neural networks. in Proceedings of the IEEE conference on computer vision and pattern recognition. 2016.
    \bibitem{15}Tashiro, Y., Y. Song, and S. Ermon, Diversity can be transferred: Output diversification for white-and black-box attacks. Advances in Neural Information Processing Systems, 2020. 33: p. 4536-4548.
    \bibitem{16}Liu, Y., et al., Practical evaluation of adversarial robustness via adaptive auto attack. arXiv preprint arXiv:2203.05154, 2022.
    \bibitem{17}Athalye, A., N. Carlini, and D. Wagner. Obfuscated gradients give a false sense of security: Circumventing defenses to adversarial examples. in International conference on machine learning. 2018. PMLR.
    \bibitem{18}Zhang, H., et al. Theoretically principled trade-off between robustness and accuracy. in International Conference on Machine Learning. 2019. PMLR.
    \bibitem{19}Jin, C. and M. Rinard, Manifold regularization for adversarial robustness. arXiv preprint arXiv:2003.04286, 2020. 1.
    \bibitem{20}Cai, Q.-Z., et al., Curriculum adversarial training. arXiv preprint arXiv:1805.04807, 2018.
    \bibitem{21}Zhang, J., et al. Attacks which do not kill training make adversarial learning stronger. in International Conference on Machine Learning. 2020. PMLR.
    \bibitem{22}Zhang, D., et al., You only propagate once: Accelerating adversarial training via maximal principle. arXiv preprint arXiv:1905.00877, 2019.
    \bibitem{23}Wong, E., L. Rice, and J.Z. Kolter, Fast is better than free: Revisiting adversarial training. arXiv preprint arXiv:2001.03994, 2020.
    \bibitem{24}Ioffe, S. and C. Szegedy. Batch normalization: Accelerating deep network training by reducing internal covariate shift. in International conference on machine learning. 2015. PMLR.
    \bibitem{25}Kingma, D.P. and J. Ba, Adam: A method for stochastic optimization. arXiv preprint arXiv:1412.6980, 2014.
    \bibitem{26}Atzmon, M., et al., Controlling neural level sets. Advances in Neural Information Processing Systems, 2019. 32.
    \bibitem{27}Andriushchenko, M. and N. Flammarion, Understanding and improving fast adversarial training. Advances in Neural Information Processing Systems, 2020. 33: p. 16048-16059.
    \bibitem{28}Sehwag, V., et al., Improving adversarial robustness using proxy distributions. arXiv preprint arXiv:2104.09425, 2021.
    \bibitem{29}Kundu, S., et al. DNR: A tunable robust pruning framework through dynamic network rewiring of DNNs. in Proceedings of the 26th Asia and South Pacific Design Automation Conference. 2021.
    \bibitem{30}Addepalli, S., et al., Towards Achieving Adversarial Robustness Beyond Perceptual Limits. 2021.
    \bibitem{31}Ding, G.W., et al., Mma training: Direct input space margin maximization through adversarial training. arXiv preprint arXiv:1812.02637, 2018.
    \bibitem{32}Engstrom, L., et al., Robustness (python library). 2019.
    \bibitem{33}Wang, Y., et al. Improving adversarial robustness requires revisiting misclassified examples. in International Conference on Learning Representations. 2019.
    \bibitem{34}Zhang, H. and J. Wang, Defense against adversarial attacks using feature scattering-based adversarial training. Advances in Neural Information Processing Systems, 2019. 32.
    \bibitem{35}Zhang, H. and W. Xu, Adversarial interpolation training: A simple approach for improving model robustness. 2019.
    \bibitem{36}Wu, D., S.-T. Xia, and Y. Wang, Adversarial weight perturbation helps robust generalization. Advances in Neural Information Processing Systems, 2020. 33: p. 2958-2969.
    \bibitem{37}Zhang, J., et al., Geometry-aware instance-reweighted adversarial training. arXiv preprint arXiv:2010.01736, 2020.
    \bibitem{38}Sehwag, V., et al., Hydra: Pruning adversarially robust neural networks. Advances in Neural Information Processing Systems, 2020. 33: p. 19655-19666.
    \bibitem{39}Hendrycks, D., K. Lee, and M. Mazeika. Using pre-training can improve model robustness and uncertainty. in International Conference on Machine Learning. 2019. PMLR.
    \bibitem{40}Carmon, Y., et al., Unlabeled data improves adversarial robustness. Advances in Neural Information Processing Systems, 2019. 32.
    \bibitem{41}Gowal, S., et al., Uncovering the limits of adversarial training against norm-bounded adversarial examples. arXiv preprint arXiv:2010.03593, 2020.
    \bibitem{42}Rebuffi, S.-A., et al., Fixing data augmentation to improve adversarial robustness. arXiv preprint arXiv:2103.01946, 2021.
    \bibitem{43}Sridhar, K., et al., Robust Learning via Persistency of Excitation. 2021.
    \bibitem{44}Huang, L., C. Zhang, and H. Zhang, Self-adaptive training: beyond empirical risk minimization. Advances in neural information processing systems, 2020. 33: p. 19365-19376.
    \bibitem{45}Kim, J. and X. Wang, Sensible adversarial learning. 2019.
    \bibitem{46}Sitawarin, C., S. Chakraborty, and D. Wagner, Improving adversarial robustness through progressive hardening. 2020.
    \bibitem{47}Cui, J., et al., Learnable Boundary Guided Adversarial Training. 2020.
    \bibitem{48}Pang, T., et al., Boosting Adversarial Training with Hypersphere Embedding. 2020.
    \bibitem{49}Rice, L., E. Wong, and Z. Kolter. Overfitting in adversarially robust deep learning. in International Conference on Machine Learning. 2020. PMLR.
\end{thebibliography}
\end{document}